\documentclass{llncs}
\usepackage{graphicx}
\begin{document}
\title{Table2answer: Read the database \newline and answer without SQL}
%
%
\author{
Tong Guo\inst{1}
Huilin Gao\inst{2}
}
%

%
\institute{Rokid AI Lab\and
China Electronic Technology Group Corporation Information Science Academy, Beijing, China}
\maketitle              
\begin{abstract}
Semantic parsing is the task of mapping natural language to logic form. In question answering, semantic parsing can be used to map the question to logic form and execute the logic form to get the answer. One key problem for semantic parsing is the hard label work. We study this problem in another way: we do not use the logic form any more. Instead we only use the schema and answer info. We think that the logic form step can be injected into the deep model. The reason why we think removing the logic form step is possible is that human can do the task without explicit logic form. We use BERT-based model and do the experiment in the WikiSQL dataset, which is a large natural language to SQL dataset. Our experimental evaluations that show that our model can achieves the baseline results in WikiSQL dataset.

\end{abstract}
\section{Introduction}
One way to construct a question-answering system over database is leveraging the semantic parsing. Semantic parsing is a task that transform the natural language to logic form which computer can execute. Transforming from natural language to SQL (NL2SQL) is kind of semantic parsing task. The generated SQL can be executed in the database system to can the answer from the database. In recent years, deep learning techniques \cite{ref_proc1} is applied to semantic parsing\cite{ref_proc4}\cite{ref_proc7}\cite{ref_proc8}\cite{ref_proc9}.
But deep learning need large amount of labeled data, when the parameter number is very large. The logic form of natural language is also very hard to label, compared to other natural language processing (NLP) tasks such as text classification or sentence similarity. There are some works\cite{ref_proc14} solve the hard labeling problem in weak supervision methods. Our work try to solve this problem in another way. The upper level view of the problem of semantic parsing for question answering is retrieve the answer from the database when given a question. Human can solve this problem even without an explicit logic form. Human can read the schema or columns' info in the database and answer the question. We think the deep model can integrate the logic or reasoning modules like \cite{ref_proc2} or other deep model to search on the database without an explicit logic form. We present our idea in Fig. 1. and Fig. 2.
\begin{figure}
\centering
\includegraphics[width=0.5\textwidth]{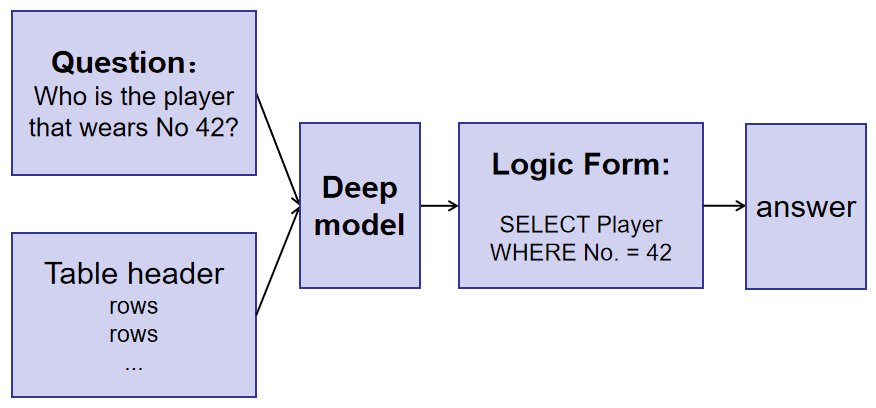}
\caption{The illustration of natural language to SQL for question answering} \label{fig1}
\end{figure}

\begin{figure}
\centering
\includegraphics[width=0.5\textwidth]{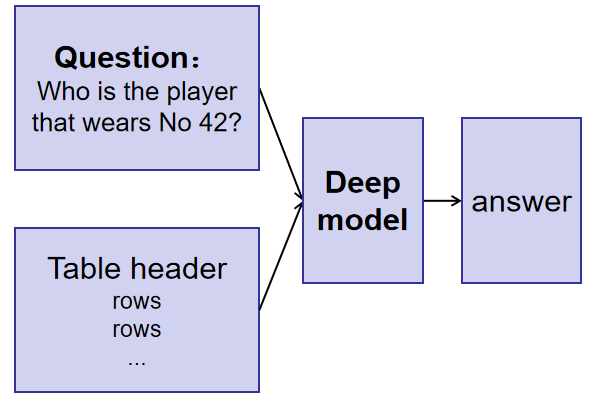}
\caption{The illustration of question answering over database without SQL} \label{fig2}
\end{figure}
We use BERT-based \cite{ref_proc3} to implement our idea. BERT
is a new language representation model, which stands for
Bidirectional Encoder Representations from Transformers.
Pre-trained word representations on a large (unlabeled) language corpus, such as \cite{ref_proc15} , have shown promising results in a lot of NLP tasks. BERT is also a pre-trained deep model\cite{ref_proc11}\cite{ref_proc12}\cite{ref_proc13} which use large amount of plain text to pre-train. As a result, the pre-trained BERT representations can be fine-tuned with just one additional output layer to create state-of-the-art models for a wide range of tasks, such as question answering and text classification.

Our main contributions in this work are two-fold. First, we 
introduce our idea of finding the answer from database with out semantic parsing to solve the problem of hard label work for semantic parsing. Second, we use BERT-based model to implement our idea and achieve a baseline level experiment result. The code is available. \footnote[1]{\url{https://github.com/guotong1988/table2answer}}

\section{Task Description}
In the NL2SQL without SQL task (Table2answer), given a question and a database table, the deep model needs to find the answer in the database table. The question is described as a sequence of word tokens: $Q = \{w_1,w_2,...,w_n\}$, where $n$ is the number of words in the question, and the table is described as a sequence of column names or headers $H=\{h_1,h_2,...,h_m\}$, where $m$ is the number of columns in the table. Each table contains a number of rows which contains the answer or cells to the question. The answer for the model is the pointers to the table cells. We denote the cells as $T = \{c_1,c_2,...,c_{r \times m}\}$, where r is the number of rows. Note that each $c_{r \times m}$ is not one word in the table, but each $c_{m \times r}$ is one cell in the table. In experiment, we concatenate the word embeddings in a cell to represent one cell. The cell representation is the input of the transformer layers of BERT.

We now describe the WikiSQL dataset \cite{ref_proc4}, a dataset of 80654 hand-annotated examples of questions and SQL queries distributed across 24241 tables from Wikipedia.  We present an example in Fig. 3. 

We extract the question, answer index and table content from the WikiSQL dataset and construct the dataset for Table2answer. One table corresponding to several questions with answers in the table. For an elementary consideration, we only extract the data case which SQL of question only contains one condition in the WHERE clause. We leave it as future work which solves all kinds of questions with SQL in the WikiSQL. We present an example in Fig. 4. Also, Our model works under the condition that the table which contains the answer is determined. In other words, our model need not predict the exact table in the all tables. We use other methods to find the exact table in industry application.
\begin{figure}
\centering
\includegraphics[width=\textwidth]{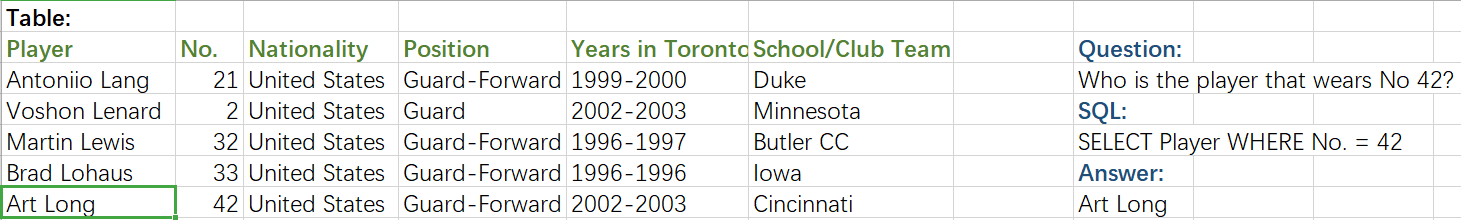}
\caption{An example of the WikiSQL semantic parsing dataset. The inputs consist of a table and a question. The outputs consist
of a ground truth SQL query and the corresponding result from execution.} \label{fig3}
\end{figure}

\begin{figure}
\centering
\includegraphics[width=\textwidth]{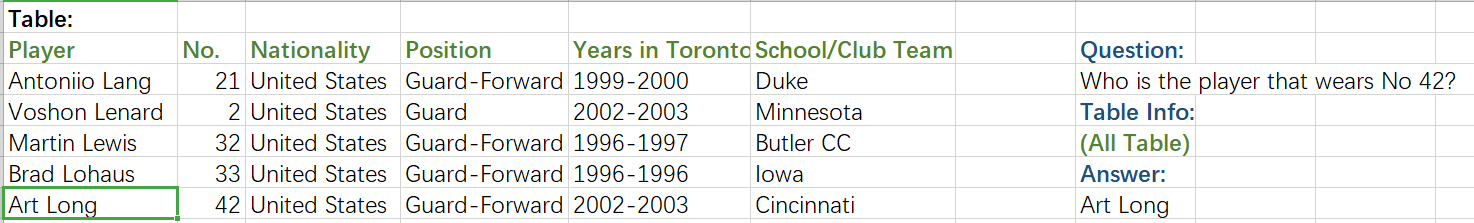}
\caption{An example of the Table2anwser dataset. The inputs consist of a table and a question. The outputs only consist the  corresponding result from SQL execution.} \label{fig4}
\end{figure}

\begin{figure}
\centering
\includegraphics[width=\textwidth]{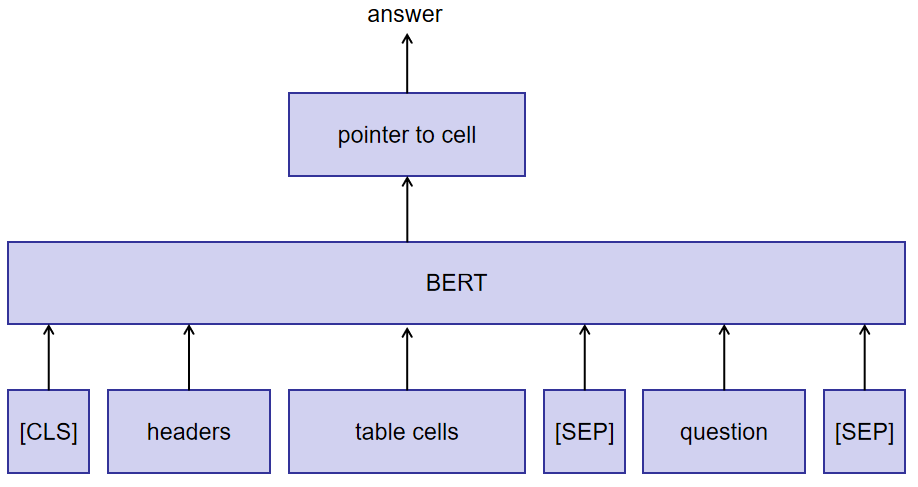}
\caption{The overall model for Table2answer} \label{fig5}
\end{figure}

\section{Model}
In this section, we describe the details of our BERT-based model to solve the problem of question answering over database without semantic parsing. We present the overall solution for the Table2answer problem in Fig. 5. The reason and inspiration to use BERT-based model is that we want to leverage the attention info between question and table header to point the exact cell in the table. 

We follow the BERT convention of data input format for encoding the natural language question together with the headers and cells of the table. We use [SEP] to separate between the cells and the question. We average word embedding of one cell for the representation of one cell. At last, each headers $H=\{h_1,h_2,...,h_m\}$, table cells $T = \{c_1,c_2,...,c_{r \times m}\}$ and question $Q = \{w_1,w_2,...,w_n\}$ is encoded as following:

$[CLS], h_1, h_2, ..., h_m, c_1, c_2, ..., c_{r \times m}, [SEP], w_1, w_2, ..., w_n, [SEP]$

In the SQuAD \cite{ref_proc5} machine comprehension task, we input the paragraph and question to the model and find the answer string in the paragraph. And our Table2answer task, we input the question, headers and table cells to find the answer cell in the table. The two tasks are very similar in this perspective. So we append one pointer \cite{ref_proc6} after the output of the BERT main module. It is different from machine comprehension task because we only have one cell to point as we have simplified the dataset. We leave it as future work to choose more than one cell as answer.

\section{Experiments}
In this section, we present more details of the model and the evaluation on the dataset. Pre-trained BERT models (BERT-Base-Uncased) are loaded and fine-tuned with Adam optimizer with learning rate $5*10^5$. The batch size is 16. We use the origin BERT tokenizer with the same vocabulary of BERT-Base-Uncased. We fix the parameters of 1-9 layers of BERT-Base and fine-tune the last 3 layers, as we observe that fine-tuning all the layers do not get a better evaluation result. Our neural network model is implemented in TensorFlow.

\subsection{Data augmentation}
We randomly shuffle the rows of all the tables and get a training dataset of 503881 data and test dataset of 1874 data. Note that the answer cell index is corresponding to the shuffled rows. We have not shuffled the columns of table as we observe bad result of it.
See  Tab. 1. for detail.
\begin{table}
\caption{The evaluation of our experiment for data augmentation.}\label{tab1}
\centering
\begin{tabular}{|l|l|}
\hline
Training Data Size & Test Accuracy\\
\hline
76301 & 20.3$\%$\\
\hline
321536 & 47.2$\%$\\
\hline
503881 & 54.0$\%$\\
\hline
\end{tabular}
\end{table}

\begin{table}
\caption{The evaluation of our experiment. Our baseline is a transformer\cite{ref_proc10} without pre-training. As there may be same answers in different cells, we consider the final word match as the accuracy.}\label{tab1}
\centering
\begin{tabular}{|l|l|}
\hline
Model & Test Accuracy\\
\hline
transformer baseline & 11.0$\%$\\
\hline
Our model & 54.0$\%$\\
\hline
Our model without data augmentation & 17.7$\%$\\
\hline
Our model without position embedding & 20.5$\%$\\
\hline
\end{tabular}
\end{table}

\subsection{Evaluation}
We evaluate our model on the dataset that extract from WikiSQL. The results are presented in Tab. 2. The training accuracy is around 96$\%$ and we leave it as future work to further improve the result on the test dataset. We also do the experiment just the same as SQuAD machine comprehension task. That is, we concatenate all the words in the table cells and append two pointer after BERT for the start index and end index. And the result for this kind of data feeding methods is 2$\%$-3$\%$ lower.

\section{Conclusion}
In this paper, in order to solve the problem that the labeling work for semantic parsing is too hard, we introduce our idea that inject the reasoning part into the deep model to remove the logic form step for question answering. We think that human can do the logic operation even without the SQL so we believe our idea will work. Then We design the BERT-based model and achieve the baseline results in the sub-WikiSQL dataset. It is trained end-to-end and can retrieve the answer directly. The dataset for table2answer is simpler than WikiSQL and there will be a lot of work to research.


\begin{thebibliography}{8}
\bibitem{ref_proc1}
A. Krizhevsky, I. Sutskever, and G. Hinton.: Imagenet classification with deep convolutional neural networks. In NIPS (2012)

\bibitem{ref_proc2}
Andrew Trask, Felix Hill, Scott Reed, Jack Rae, Chris Dyer, Phil Blunsom: Neural Arithmetic Logic Units. 
arxiv.org/abs/1808.00508

\bibitem{ref_proc3}
Jacob Devlin, Ming-Wei Chang, Kenton Lee, and Kristina Toutanova. 2018. BERT: pre-training of deep bidirectional transformers for language understanding. CoRR, abs/1810.04805.

\bibitem{ref_proc4}
Victor Zhong, C. Xiong, and R. Socher. Seq2SQL: Generating Structured Queries from Natural Language using Reinforcement Learning. arXiv preprint arxiv:1709.00103, Nov 2017

\bibitem{ref_proc5}
Rajpurkar P, Zhang J, Lopyrev K, et al: Squad: 100,000+ questions for machine comprehension of text[J]. arXiv pre-print arXiv:1606.05250 (2016)

\bibitem{ref_proc6}
Oriol Vinyals, Meire Fortunato, and Navdeep Jaitly: Pointer networks. International Conference on Neural Information Processing Systems. MIT Press (2015)

\bibitem{ref_proc7}
Xiaojun Xu, Chang Liu, Dawn Song. 2017. SQLNet: Generating Structured Queries from Natural Language Without Reinforcement Learning.

\bibitem{ref_proc8}
Yu T, Li Z, Zhang Z, et al. TypeSQL: Knowledge-based Type-Aware Neural Text-to-SQL Generation[J]. arXiv preprint arXiv:1804.09769, 2018.

\bibitem{ref_proc9}
Dong, Li, and Mirella Lapata. "Coarse-to-Fine Decoding for Neural Semantic Parsing." arXiv preprint arXiv:1805.04793 (2018).

\bibitem{ref_proc10}
Ashish Vaswani, Noam Shazeer, Niki Parmar, Jakob Uszkoreit, Llion Jones, Aidan N. Gomez, Lukasz Kaiser, and Illia Polosukhin. 2017. Attention is all you need. CoRR, abs/1706.03762

\bibitem{ref_proc11}
Matthew Peters, Mark Neumann, Mohit Iyyer, Matt
Gardner, Christopher Clark, Kenton Lee, and Luke
Zettlemoyer. 2018. Deep contextualized word representations. In NAACL.

\bibitem{ref_proc12}
Alec Radford, Karthik Narasimhan, Tim Salimans, and
Ilya Sutskever. 2018. Improving language understanding with unsupervised learning. Technical report, OpenAI.

\bibitem{ref_proc13}
Andrew M Dai and Quoc V Le. 2015. Semi-supervised
sequence learning. In Advances in neural information processing systems, pages 3079-3087

\bibitem{ref_proc14}
Liang C, Berant J, Le Q, et al. Neural symbolic machines: Learning semantic parsers on freebase with weak supervision[J]. arXiv preprint arXiv:1611.00020, 2016.

\bibitem{ref_proc15}
Jeffrey Pennington, Richard Socher, and Christopher
D. Manning. 2014. Glove: Global vectors for word representation. In EMNLP.

\end{thebibliography}
\end{document}